\documentclass[sigconf,anonymous=false]{acmart}

\AtBeginDocument{%
  \providecommand\BibTeX{{%
    \normalfont B\kern-0.5em{\scshape i\kern-0.25em b}\kern-0.8em\TeX}}}

\copyrightyear{2022}
\acmYear{2022}
\setcopyright{acmcopyright}
\acmConference[MM '22]{Proceedings of the 30th ACM International Conference on Multimedia}{October 10--14, 2022}{Virtual Event, Portugal}
\acmBooktitle{Proceedings of the 30th ACM International Conference on Multimedia (MM '22), October 10--14, 2022, Virtual Event, Portugal}
\acmPrice{15.00}
\acmDOI{10.1145/3474085.xxxxxxx}
\acmISBN{978-1-4503-8651-7/22/10}

\usepackage{multirow}
\usepackage{amsmath}
\usepackage{algorithm}
\usepackage{algorithmic}
\usepackage{caption}
\usepackage{subfigure}
\DeclareMathOperator*{\argmax}{arg\,max}

\begin{document}
\fancyhead{}
\title{Towards Visual-Prompt Temporal Answering Grounding  in Medical Instructional Video}

\author{Bin Li$^{1}$*,\quad Yixuan Weng$^{2}$*$^{\dagger}$,\quad Bin Sun$^{1}$,\quad Shutao Li$^{1}$$^{\ddagger}$}

\makeatletter
\def\authornotetext#1{
\if@ACM@anonymous\else
    \g@addto@macro\@authornotes{
    \stepcounter{footnote}\footnotetext{#1}}
\fi}
\makeatother
\authornotetext{These authors contribute equally to this work.}
\authornotetext{Work done during an internship at Chinese Academy Sciences.}
\authornotetext{Corresponding author.}
\affiliation{
 \institution{\textsuperscript{\rm 1}College of Electrical and Information Engineering, Hunan University, Changsha, China}
 \institution{\textsuperscript{\rm 2}National Laboratory of Pattern Recognition Institute of Automation,
 	Chinese Academy Sciences, Beijing, China}
 \country{}
 }
 
\email{{libincn, sunbin611, shutao\_li}@hnu.edu.com}
\email{wengsyx@gmail.com}

\def\authors{Anonymous Author(s)}
\renewcommand{\shortauthors}{Anonymous Author(s)}
\begin{abstract}
	The temporal answering grounding in the video (TAGV) is a new task naturally derived from temporal sentence grounding in the video (TSGV). Given an untrimmed video and a text question, this task aims at locating the matching span from the video that can semantically answer the question. Existing methods tend to formulate the TAGV task with a visual span-based question answering (QA) approach by matching the visual frame span queried by the text question. However, due to the weak correlations and huge gaps of the semantic features between the textual question and visual answer, existing methods adopting visual span predictor perform poorly in the TAGV task. To bridge these gaps, we propose a visual-prompt text span localizing (VPTSL) method, which introduces the timestamped subtitles as a passage to perform the text span localization for the input text question, and prompts the visual highlight features into the pre-trained language model (PLM) for enhancing the joint semantic representations. Specifically, the context query attention is utilized to perform cross-modal interaction between the extracted textual and visual features. Then, the highlight features are obtained through the video-text highlighting for the visual prompt. To alleviate semantic differences between textual and visual features, we design the text span predictor by encoding the question, the subtitles, and the prompted visual highlight features with the PLM. As a result, the TAGV task is formulated to predict the span of subtitles matching the visual answer. Extensive experiments on the medical instructional dataset, namely MedVidQA, show that the proposed VPTSL outperforms the state-of-the-art (SOTA) method by 28.36\% in terms of mIOU with a large margin, which demonstrates the effectiveness of the proposed visual prompt and the text span predictor.
\end{abstract}




\maketitle
\section{Introduction}
\begin{figure}[h]
	\centering
	\includegraphics[width=1.03\linewidth]{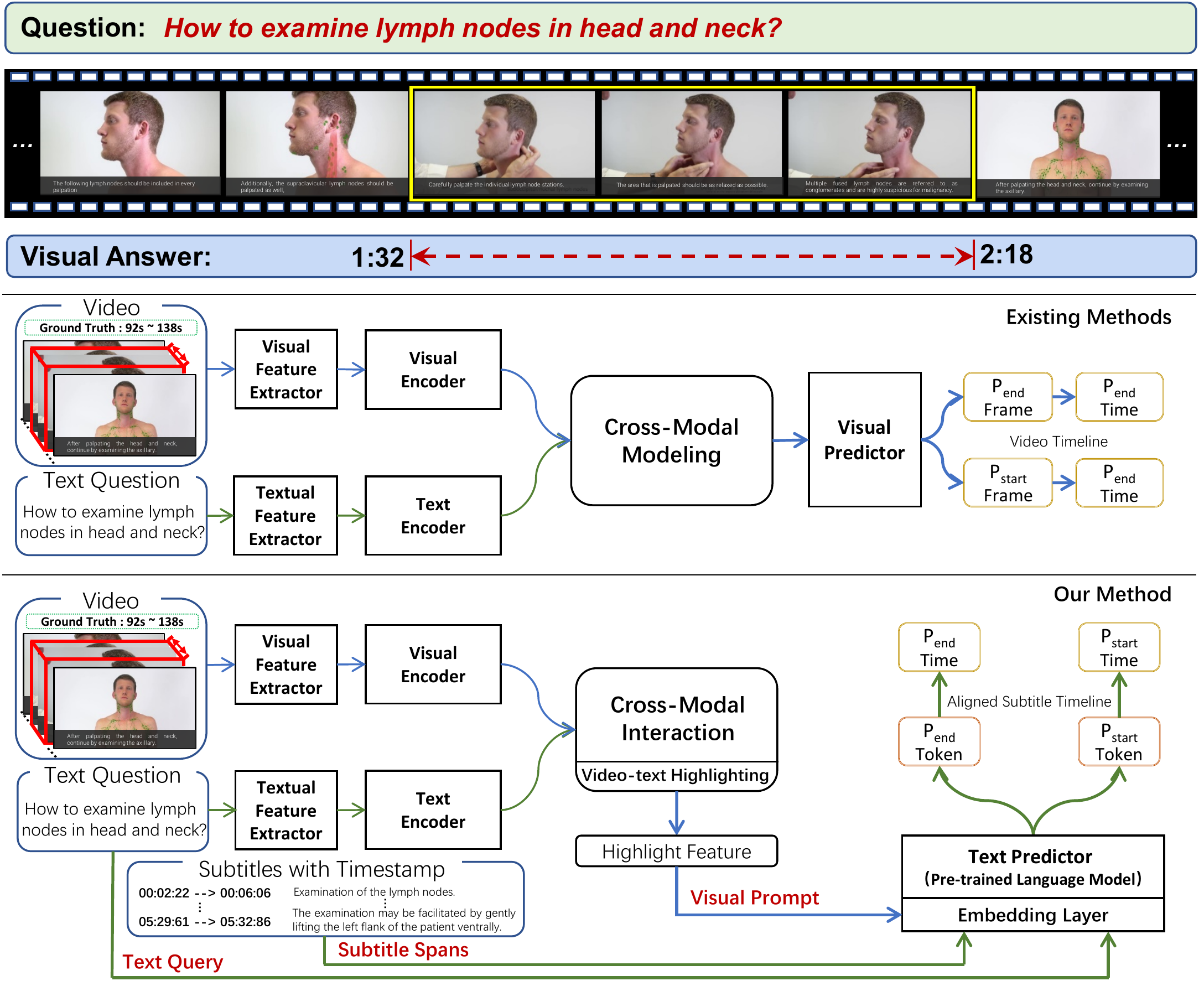}
	\caption{Illustration of the temporal answering localization in the medical instructional video, where the visual answer with the subtitles locates in the video timeline to perform a demonstration. Below are the differences between the existing method and our method.
	}
	\vspace{0.1cm}
	\label{sample2}
\end{figure}
``Hey, Siri, could you please show me how to examine lymph nodes in the head and neck ?'' Then, the video containing the right processes comes into our eyes... Recently, the surge in availability of online videos has changed the way of acquiring information and knowledge \cite{monfort2021spoken, shelke2021comprehensive, khan2021transformers}. Many people prefer instructional videos to teach or learn how to accomplish a particular task with a series of step-by-step procedures \cite{ghoddoosian2022hierarchical}. The temporal answering grounding in the video (TAGV) is a new task that has attracted increasing attention due to the visual and verbal communication at the same time in an effective and efficient manner \cite{MedVidQA2022, gupta2022dataset}. The goal of the TAGV task is to find the matching video answer span corresponding to its question, aka., visual answering localization. As the natural development from temporal sentence grounding in the video (TSGV) \cite{yang2020survey, zhang2022elements}, the TAGV task is challenging since there are huge gaps between two different modalities. The text is discontinuous in syntactic structure, while the video is continuous within adjacent frames \cite{chen2018temporally}. People can easily answer through natural language but are hard to act without the moment guidance in the video to demonstrate their answers. As shown at the top of Figure \ref{sample2}, this example illustrates the temporal answering localization in the medical instructional video,  where the figure is borrowed from the original work \cite{gupta2022dataset} with the author's permission and change. As we can see in this figure, the particular temporal answering segment is preferred rather than the entire video as the answer to the given question "How to examine lymph nodes in a head and neck ?". How to design a cross-modal method that can locate the video timeline correctly is still one of the key points in the current research \cite{gupta2022dataset,zhang2020span}. \par
Many efforts have been made to realize a reliable and accurate natural language temporal localization in the video \cite{zhang2020span, xiao2021natural, zhang2021natural}, where similar tasks are proven to be important for cross-modal understanding, such as video moment retrieval (VMR)\cite{9374685}, and video question answering (VQA)\cite{lei2018tvqa}. On the query side, the query of the TAGV task is a text question instead of a direct text description in the VMR. On the answer side, the answer of the TAGV is located on the video timeline different from the text answering the visual question in the VQA. Therefore, the existing methods may perform poorly in the TAGV task. Similar to the question answering (QA) problem in the  natural language processing (NLP) field, we resort to the existing span-based grounding methods \cite{zhang2020span, zhang2021natural} to address the TAGV problem. \par
As shown in the middle of Figure \ref{sample2}, existing span-based methods tend to encode video and text separately for feature encoding and adopt cross-modal modeling to construct feature representations in the same space. The visual answer spans can be located with the head and tail in the video frame. However, there is a huge difference in the semantic information between text and video \cite{opazo2019proposal, HaoyuTang2021FramewiseCM},  where the located video spans queried by the text question may be biased. Moreover, the weak correlations between text queries and video frames may lead to insufficient representation for an answer \cite{xu2022transformers}. \par
To address the above issues, we propose a visual-prompt text span localization (VPTSL) method, which aims to adopt visual highlight features to enhance the text span localization with the pre-trained language model (PLM). Different from the existing methods, we leverage the highlight feature as the visual prompt to enhance textual features from the PLM,  where the joint semantic representations can be learned together. Given a text question, the timestamped subtitle to the visual answer is modeled to be predicted as the final result. We illustrate the proposed VPTSL method at the bottom of Figure \ref{sample2}.  \par
Our main contributions are three-fold:

\begin{itemize}
\item To the best of our knowledge, this is the very first attempt to apply the text span predictor for solving the temporal answering grounding problem, where the timestamps of the subtitle corresponding to the visual answer are formulated for prediction.
\item The visual highlight features are designed to prompt the visual information for the textual features, where the verbal and the visual part of the video can be jointly learned through the PLM. 
\item Extensive experiments are performed to demonstrate the effectiveness of the proposed VPTSL on the medical instructional dataset (MedVidQA), in which we achieve 28.36 in mIOU score by a large margin compared with other state-of-the-art methods.
\end{itemize}
\section{Related Work}
\subsection{Temporal Sentence Grounding in Video }
The temporal sentence grounding in the video  (TSGV) is a critical task for cross-modal understanding \cite{9262795, zhang2022elements}. This task takes a video-query pair as input where the video is a collection of consecutive image frames and the query is a sequence of words. Early attempts resort to the sliding window-based \cite{anne2017localizing, gao2017tall, MengLiu2018AttentiveMR} and scanning-and-ranking based \cite{chen-etal-2018-temporally, ge2019mac, zhang2019man, liu2020jointly} paradigm. The former first generates multiple segments and then ranks them according to the similarity between segments and the query. The latter samples candidate segments via the sliding window mechanism and subsequently integrates the query with each segment representation via a matrix operation. The latest works tend to model this problem without segment proposal, which predicts answers directly without generating candidate answers \cite{HaoZhang2020SpanbasedLN}. For this convenience, many works tend to adopt the visual span predictor for locating the sentence grounding segments, where more efficient cross-modal interaction modeling are designed \cite{HaoyuTang2021FramewiseCM, zhang2020span, zhang2021natural}. However, due to the gaps between the textual features and visual features \cite{10.1145/3394171.3414053, 10.1145/3394171.3413840}, current methods adopted in the TSGV performs poorly in the temporal answering grounding in the video (TAGV). Different from them, our method tries to model subtitles with timestamps for locating the visual answer. The text span predictor is designed in the proposed method, where more semantic information between the predicted answers and the input text question can be jointly learned through the pre-trained language model.
\begin{figure*}[t]
	\centering
	\includegraphics[width=16cm]{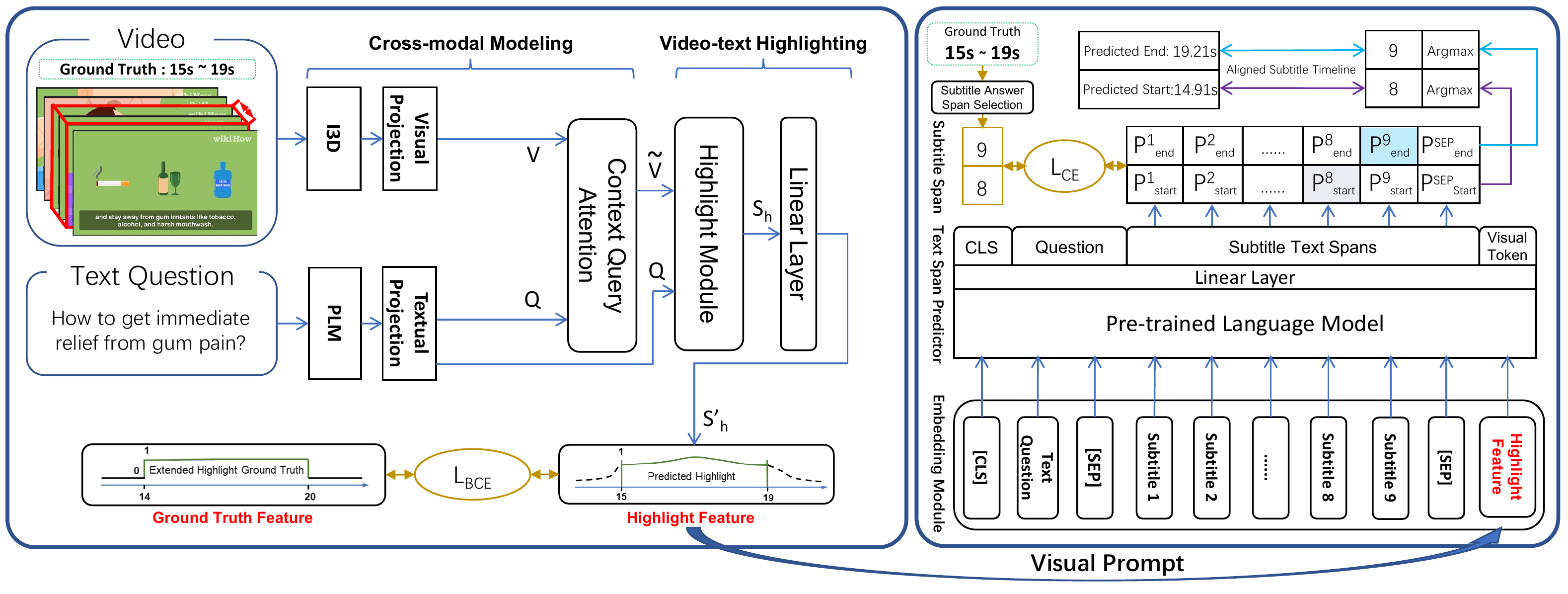}
	\caption{Overview of the proposed Visual-Prompt Text Span Localization (VPTSL) method.}
	\label{framework}
\end{figure*}

\subsection{Prompt Learning Tuning}
The concept of prompt tuning originates from the NLP domain \cite{liu2021pre}, whose  motivation is to provide pre-trained language models, such as BERT~\cite{devlin2019bert} or GPT~\cite{radford2019language}, with extra knowledge. Specifically, given a pre-trained language model, the manual designed templates are used to augment the input with extra information \cite{reynolds2021prompt}. The basic idea of prompting is to induce a pre-trained language model for downstream prediction given cloze-style prompts, such as sentiment analysis \cite{wei2021pretrained}. The key lies in how to design the prompt part for tuning the pre-trained model \cite{lester2021power}. In the computer vision field, prompt learning is a nascent research direction that has only been explored very recently \cite{yao2021cpt, rao2021denseclip, zhou2021learning}. The pioneering works have designed many efficient modules of cross-modal interaction for the downstream tasks  \cite{liu2022things, zhou2022conditional}, where the features of different modalities are optimised continuously in the embedding space.
Our method is based on the pre-trained language model, adopting the visual prompt feature for perceiving the verbal and non-verbal parts. Concretely, the video-text highlighting is designed for capturing the frame supervision for the text span predictor. The visual prompt features are utilized as the visual tokens for enhancing the pre-trained language model with non-verbal semantics.
\begin{figure}[t]
	\centering
	\includegraphics[width=8.5cm]{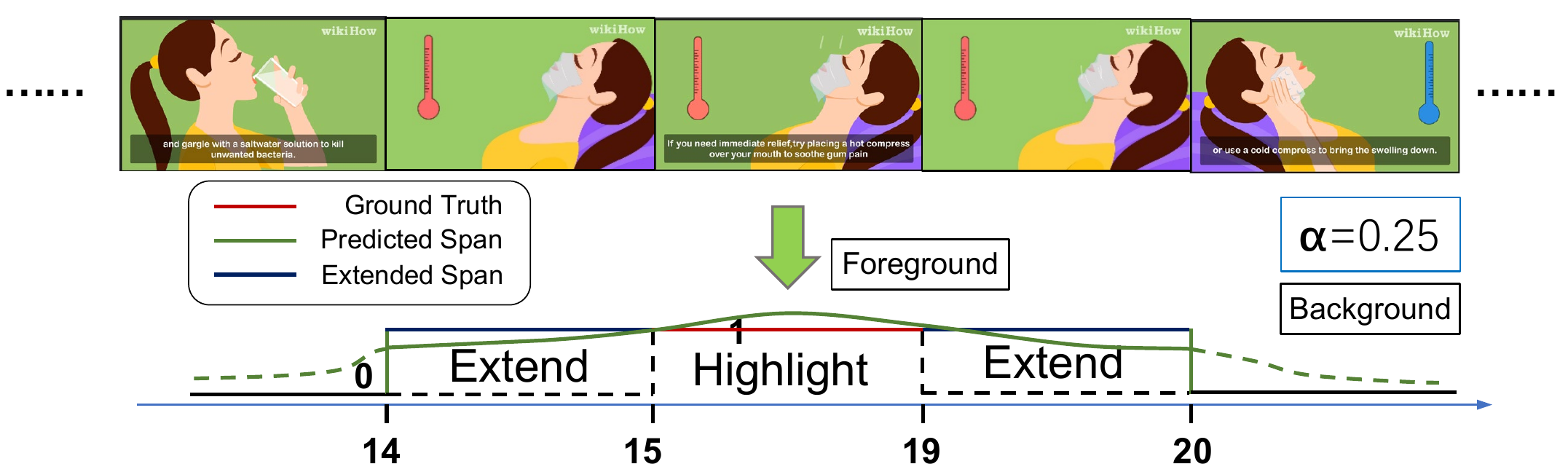}
	\caption{The illustration of visual highlighting.}
	\label{high}
\end{figure}
\section{Main Method}
\label{a1}
We propose the Visual-Prompt Text Span Localization (VPTSL) method for the TAGV task, whose goal is to predict the span of the subtitle timestamp matching the answering frame timeline with the pre-trained language model. The overview of VPTSL is illustrated in Figure \ref{framework}, which consists of four components: (1) cross-modal modeling: the extracted visual and the textual features are processed through the cross-modal interaction. (2) Video-text highlighting: the text question is used to query the video frames for obtaining the predicted highlight feature supervised by the highlight ground truth. (3) Visual prompt: highlight feature is adopted to prompt the pre-trained language model, where the textual features can capture the visual information while jointly learning. (4) Text span predictor: the textual tokens with highlight prompts are encoded through the pre-trained language model to predict the subtitle timestamp spans. 
\subsection{Cross-modal Modeling}
\label{sec:cross}
Given an untrimmed video as $V=\{f_t\}_{t=1}^{T}$ and the text question as $Q=\{q_j\}_{j=1}^{k}$, where $T$ and $k$ are the number of frames and tokens, respectively. For obtaining the well-formed semantic representations of two modalities, we will elaborate on feature extractor and cross-modal interaction.
\subsubsection{Feature Extractor}
For each video $V$, we extract frames (16 frames per second) and then obtain the corresponding RGB visual features $\mathbf{V}^{\prime}=\{\mathbf{v}_i\}_{i=1}^{n}  \in \mathbb{R}^{n \times d_v}$ using 3D ConvNet~ (I3D) pre-trained on the Kinetics dataset \cite{JoaoCarreira2017QuoVA}, where $n$ is the number of extracted features and $d_v$ is the dimension of the visual features. The extracted features is sent to a visual projection for obtaining the visual feature $\mathbf{V} \in \mathbb{R}^{n \times d_v}$. The visual projection is designed as the Conv1D \cite{ma2021end} module with dropout (p=0.1). For the text question part, we tokenize the question into the tokens with the tokenizer. Then, the textual tokens are encoded through the DeBEATa pre-trained language model \cite{he2021debertav3} for obtaining the well-formed textual features $\{w_1, w_2, \ldots, w_m\} \in \mathbb{R}^{m \times d_w} $, where the $m$ is the length of text question and the $d_w$ is the dimension of output encoding. After performing the linear projection, the final textual features $\mathbf{Q} \in \mathbb{R}^{m \times d_t}$ is obtained, where $d_t = d _v$.  
\begin{figure}[t]
	\centering
	\includegraphics[width=0.42\textwidth]{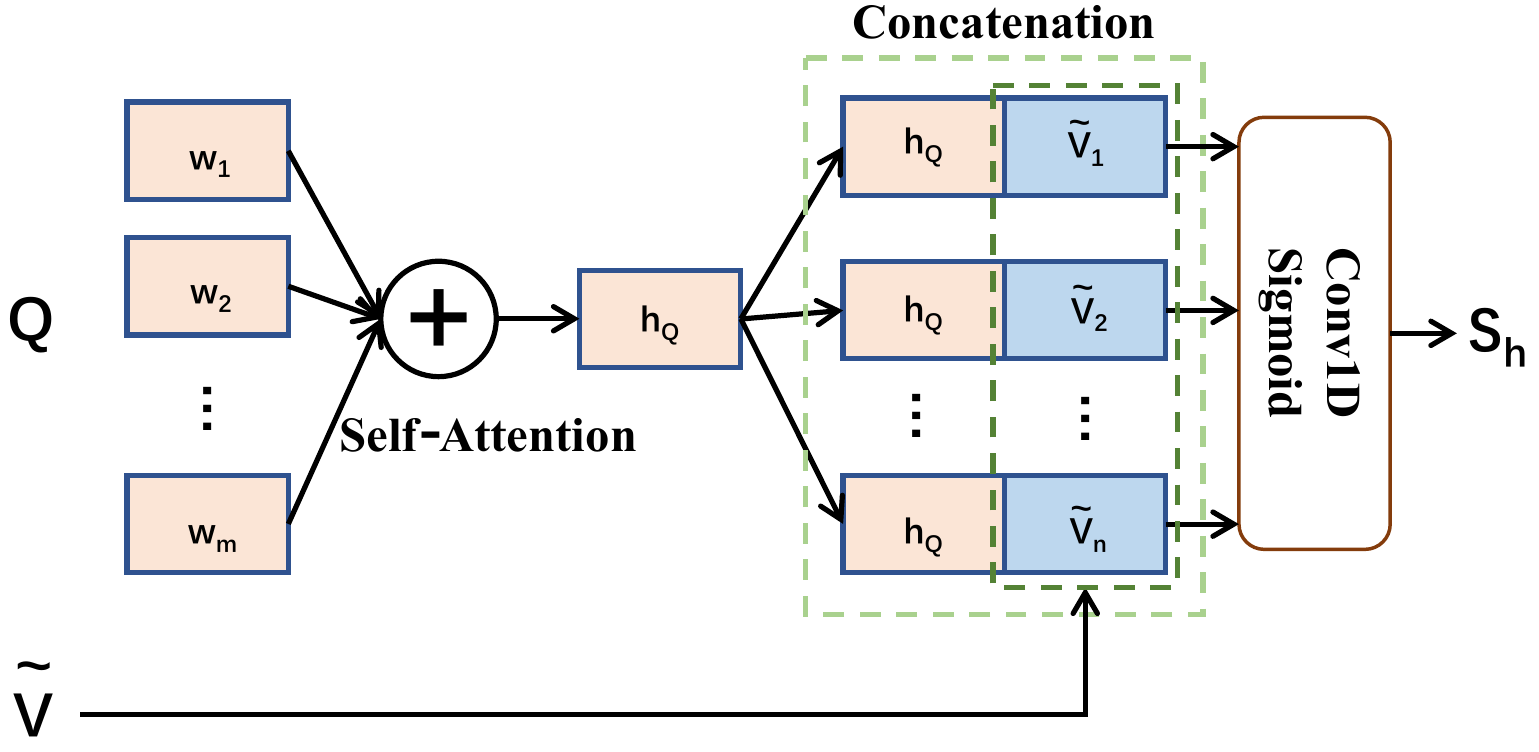}
	\caption{\small The structure of Highlighting module.}
	\label{fig_qgm}
\end{figure}
\subsubsection{Cross-modal Interaction}
After obtaining both the visual ($\mathbf{V}$) and textual ($\mathbf{Q}$) features, we perform the Context Query Attention, which is inspired by the work \cite{zhang2020span}. This module aims to capture the cross-modal interactions through context-to-query  ($\mathcal{A}$) and query-to-context ($\mathcal{B}$) process. The attention weights are computed as:\begin{equation}
	\mathcal{A}=\mathcal{S}_{r}\cdot\mathbf{{Q}}\in\mathbb{R}^{n\times d_t}, \mathcal{B}=\mathcal{S}_{r}\cdot\mathcal{S}_{c}^{T}\cdot\mathbf{{V}}\in\mathbb{R}^{n\times d_v}\nonumber
\end{equation}
where $\mathcal{S}_{r}$ and $\mathcal{S}_{c}$ are the row-wise and column-wise normalization of $\mathcal{S}$ by SoftMax, respectively. Finally, the output of context-query attention is written as:
\begin{equation}
	\mathbf{\widetilde V}={FFN}\big([\mathbf{{V}};\mathcal{A};\mathbf{{V}}\odot\mathcal{A};\mathbf{{V}}\odot\mathcal{B}]\big)
\end{equation}
where the ${FFN}$ is a single feed-forward layer, and $\odot$ denotes element-wise multiplication.
\subsection{Video-text Highlighting}
\subsubsection{Highlight Module}
\label{vh}
Inspired by the work \cite{zhang2020span}, we design the visual highlight module, which aims to percept the non-verbal part in the videos. As shown in Figure \ref{high}, the ground truth span locates in the verbal part, where the subtitles are contained. However, for an instructional video, the non-verbal part also counts a lot, so the highlight module is designed to enlarge the ground truth of text span. Specifically, we consider the verbal part as the foreground and the rest are the background in the video.  
The target text span boundaries are enlarged to cover the verbal and the non-verbal information, where the extension ratio is controlled by the hyperparameter $\alpha$.
The highlight ground truth time span is calculated as follows
\begin{equation}
	{T_{Highlight}}=t^e - t^s,
\end{equation}
where ${T_{Highlight}}$ is the highlight ground truth time span, the $t^e$ is the end ground truth time, while the $t^s$ is the start ground truth time. \par
Similar to the work \cite{zhang2020span}, we extend the non-verbal frames in the video as the extend part, which can be calculated as 
\begin{equation}
	{T_{extend}}={T_{Highlight}} * (\alpha + 1)
\end{equation}
where ${T_{extend}}$ is the extend highlight ground truth time, the $\alpha$ is the hyperparameter.
\par
The textual features into the Highlight Module are denoted as $\mathbf{{Q}}$, where the $\mathbf{{Q}}= [{{w_1}}, {{w_2}}\ldots{{w_m}} ] \in \mathbb{R}^{m \times d_t} $. The self-attention mechanism \cite{bahdanau2014neural} is performed to obtain the textual features $\mathbf{{h}_Q}\in \mathbb{R}^{1 \times d_t}$.
Then ${h}_Q$ is concatenated with each feature in $\mathbf{\widetilde{V}}$ as $\mathbf{\mathbf{\widetilde{V}}}=[\mathbf{\widetilde{v}}_{1},\dots,\mathbf{\widetilde{v}}_{n}] \in \mathbb{R}^{n \times d_v}$, where $\mathbf{\widetilde{v}}_{i}^{q}=[\mathbf {{h}_Q}; \mathbf{\widetilde v}_{i}]$, $i \in [1,n]$.
The highlighting score is computed as:
\begin{equation}
	\mathcal{S}_\textrm{h}=\sigma\big(\mathtt{Conv1D}(\mathbf{\widetilde{v}}_{i}^{q})\big)\nonumber
\end{equation}
where $\sigma$ denotes Sigmoid activation, $\mathcal{S}_{\textrm{h}}\in\mathbb{R}^{n}$. 
\subsubsection{Highlight Projection}
\par The highlighted features are required to be projected to the same dimension with the textual feature, which can be calculated by:
\begin{equation}
	\mathbf{\mathcal{S}^{\prime}_\textrm{h}}=Linear(\mathcal{S}_\textrm{h})
\end{equation}
\subsubsection{Highlight Optimization}
Accordingly, the highlight loss is computed with the BCE loss function, which is formulated as:
\begin{equation}
	\mathcal{L}_{\textrm{highlight}}=f_{\textrm{BCE}}(\mathcal{S}^{\prime}_\textrm{h}, T_{extend})
\end{equation}
\par
Moreover, the highlight module is trained in an end-to-end manner, where one of the total loss can be written as the $\mathcal{L}_1$, which is shown as follows
\begin{equation}
	\mathcal{L}_1=\mathcal{L}_{\textrm{highlight}}.
\end{equation}
\subsection{Visual Prompt}
\subsubsection{Prompt Designing} 
We use the visual highlight features as the visual token for prompting the pre-trained language model. Specifically, the highlight feature has the same dimension as the input text tokens, which is considered to be the visual token. On the one head, the visual prompt covers the non-verbal part that the text token may lack. On the other head, the visual prompt is supervised by the visual frames, where some visual features can provide the extra information as the knowledge for the pre-trained model when prompt tuning \cite{zhou2021learning}.
\subsubsection{Prompt Tuning} 
Prompt tuning is considered to be a wise choice to enhance the pre-trained model with extra knowledge \cite{liu2021pre, yao2021cpt}. Intuitively, the prompt feature is used as the visual token which concatenates with the text query (question) and the video subtitles. The [CLS] is placed at the head of the input token, while the [SEP] is used as the separator. After concatenation, each subtitle is segmented by the subtitle span, which is used for text span prediction. Then the embedding module is adopted for learning the textual and visual features jointly in the same vector space.

\subsection{Text Span Predictor}
The text span predictor is designed to predict the subtitle answer span corresponded to its visual answers. In this section, we first elaborate on the subtitle answer span selection algorithm for selecting the most proper subtitle answer span. Then, the subtitle span prediction is introduced for obtaining the final subtitle timeline.

\subsubsection{Subtitle Answer Span Selection}
\label{sass}
\begin{algorithm} [t]
	\caption{Subtitle Answer Span Selection} 
	\begin{algorithmic}
		\renewcommand{\algorithmicrequire}{\textbf{Input:}}
		
		\renewcommand{\algorithmicensure}{\textbf{Output:}}
		
		\REQUIRE Subtitle collections $D$ with time stamp, where each subtitle has its corresponding  timestamp ($d_{start}$, $d_{end}$); Start time of the visual answer $\tau_s$; End time of visual answer $\tau_e$; 
		\ENSURE  Subtitle start and end time \textbf{($R_S$, $R_E$)}
		
		\STATE $TimeStart_{min}$ $\gets+\infty$ 
		
		\STATE $TimeEnd_{min}$ $\gets +\infty$ 
		
		\FOR{$i \in D$ }
		\IF{$\lvert i.d_{start}-\tau_s$  $ \rvert  <TimeStart_{min}$}
		\STATE $TimeStart_{min}$ $ \gets $  $  \lvert$ $i.d_{start}-\tau_s \rvert $ 
		
		\STATE  $R_S$ $\gets$ $i.d_{start}$
		\ENDIF
		\IF{$\lvert$$i.d_{end}-\tau_e$$\rvert$ <= $TimeEnd_{min}$}
		\STATE $TimeEnd_{min}$ $\gets$ $\lvert$$i.d_{end} - \tau_e$$\rvert$
		
		\STATE  $R_E$ $\gets$ $i.d_{end}$
		\ENDIF
		\ENDFOR
	\end{algorithmic} 
	\label{algorithm1}
\end{algorithm}
Subtitle answer span selection aims to select the most approaching text subtitle span corresponding to its visual answer. As a result, we design the aligned subtitle answer span selection for further text span prediction.  As shown in the Algorithm \ref{algorithm1}, we use the subtitle collections $D$ of the video to locate the most approaching start and end time of the visual answers $(T_S, T_E)$, It is noted that the algorithm has its limitation as the selected part of the subtitle may be inaccurate. A precise subtitle timeline location may improve the final prediction performance. We leave these to the future work.
\subsubsection{Subtitle Span Prediction} 
\label{sec:videopre}
The $\tau^{s}$ and $\tau^{e}$ represent the start and end time of the temporal moment that answers the question. Different from the visual span proposed by work \cite{zhang2020span, zhang2021natural}, we formulate the span timeline prediction as finding its corresponding subtitle timestamp. This problem is formulated as the SQuAD \cite{rajpurkar2016squad} style triples (Context, Question, Answer), where the more efficient method to locate subtitles span can be designed. As a result, we design a text span predictor based on a pre-trained language model. The input is appended with the visual prompt method, where the textual and visual tokens are learned jointly in the pre-trained model. Specifically, we use DeBERTa for feature encoding. Each token is segmented in the subtitle span, which has a probability of being selected head and tail. Therefore, subtitle span-based prediction can be performed by the cross-entropy optimization token by token. 
\par As shown in Figure \ref{framework}, the ground truth visual timeline is (15 \textasciitilde ~ 19). This frame timeline can be translated into the subtitle span stamp, which locates in spans 8 and 9. The predicted start index shown in the Figure \ref{framework} is located in the $P_{start}^{8}$, while the predicted end index locates in the $P_{end}^{9}$. So the corresponding aligned subtitle stamp can be used as the final results (14.91 \textasciitilde ~ 19.21). It is noticed that the token-level segment may present errors between the subtitle span timeline and the ground truth span timeline. As mentioned in the section \ref{sass}, more precise subtitle timeline selection may bring more accuracy for the final results. Next, we will introduce the details of subtitle span prediction.
\par
Let $\text{DeBERTa}(\cdot)$ be the pre-trained model, we first obtain the hidden representation with
\begin{equation}
	 h=\text{DeBERTa}(x) \in \mathbb{R}^{r_h*|x|}
\end{equation}
where $|x|$ is the length of the input sequence and $r_h$ is the size of the hidden dimension. 
\par Then the hidden representation is passed to two separate dense layers followed by softmax functions: 
\begin{equation}
l_1=\text{softmax}(W_1 \cdot h + b_1)
\end{equation}
\begin{equation}
l_2=\text{softmax}(W_2 \cdot h + b_2)
\end{equation}
where $W_1$, $W_2 \in \mathbb{R}^{r_h}$ and $b_1, b_2 \in \mathbb{R}$. The \text{softmax} is applied along the dimension of the sequence.
\par
The output is a span across the positions in $d$, indicated by two pointers (indexes) $s$ and $e$ computed from $l_1$ and $l_2$: 
\begin{equation}
s=\argmax_{start}(l_1)
\label{l1}
\end{equation}
\begin{equation} 
e=\argmax_{end}(l_2)
\label{l2}
\end{equation}
where equation (\ref{l1}) represents the start token of the start span, while the equation (\ref{l2}) shows the end token of the end span. 
\par  In the end, the final visual answer span will always be aligned with the text span predicted text span, which is presented as $(start $\textasciitilde ~ \\ $ end)$. The span prediction loss is optimized by minimizing the following loss:
\begin{equation}
	\mathcal{L}_2=\mathcal{L}_{\textrm{text\_span}}
\end{equation}
\subsection{Training and Inference}
\label{sec:trainandinf}
\subsubsection{Training.} 
The total optimizing function is performed as multi-loss form, which is presented as follows.
\begin{equation}
  \mathcal{L}_{total} = \lambda * \mathcal{L}_{\textrm{highlight}} +  \mathcal{L}_{\textrm{text\_span}}
\end{equation}
where the $\lambda$ is the hyper-parametor for tuning the total loss, the $ \mathcal{L}_{\textrm{highlight}}$ part provides the non-verbal information and the loss $ \mathcal{L}_{\textrm{text\_span}}$ covers the verbal text information.
\subsubsection{Inference.}  We simply take the visual highlight feature to prompt the pre-trained language model, aiming at covering the non-verbal information for the text span localization. The text span predictor performs prediction after encoding the text tokens and the visual token by the pre-trained language model. The predicted start token locates the start span, while the predicted end token locates the end span.
\section{Experiments}
In this section, we first introduce the dataset used in the experiments. Then, we elaborate on the evaluation metrics and describe the compared state-of-the-art methods. Finally, we present the implementation details.
\subsection{Datasets}
Medical Video Question Answering ({MedVidQA}) datasets \cite{gupta2022dataset} is the first video question answering (VQA) dataset \cite{PengchuanZhang2021VinVLRV} constructed in natural language video localization (NLVL)  \cite{LisaAnneHendricks2017LocalizingMI,MengLiu2018AttentiveMR} , which aims to provide medical instructional video with text question query. Three medical informatics experts were asked to formulate the medical and health-related instructional questions by watching the given video. They were required to localize the visual answer to those instructional questions by providing their timestamps in the video. 
\par
The MedVidQA dataset is composed of 899 videos with 3010 questions and the corresponding visual answers. The mean duration time of these videos is 383.29 seconds. The MedVidQA provides subtitle information of the original video and visual feature information extracted from the 3D ConvNet (I3D) which was pre-trained on the Kinetics dataset \cite{JoaoCarreira2017QuoVA}. We follow the official data split, where 2710, 145, and 155 questions and visual answers are used for training, validation, and testing respectively. 
\subsection{Evaluation Metrics} 
Following prior works \cite{JiyangGao2017TALLTA, MengLiu2018AttentiveMR,YitianYuan2018ToFW,HaoZhang2020SpanbasedLN,gupta2022dataset}, we adopt ``R@n, IoU = $\mu$'' and ``mIoU'' as the evaluation metrics, which treats localization of the frames in the video as a span prediction task similar to answer span prediction \cite{WenhuiWang2017GatedSN, MinjoonSeo2016BidirectionalAF} in text-based question answering. The ``R@n, IoU = $\mu$'' denotes the percentage of language queries having at least one result whose Inter-section over Union (IoU) with ground truth is larger than $\mu$ in top-n retrieved moments. ``mIoU'' is the average IoU over all testing samples. In our experiments, we use n = 1 and $\mu \in {0.3,0.5,0.7}$. The calculation equation is shown as follows
\begin{equation}
	mIOU = ({\sum_{i=1}^{n} \frac{A_n \cap B_n}{A_n \cup B_n}})/{n}
\end{equation}
where $A$ and $B$ represent different span collections.
\begin{table*}[t]
	\centering
	\renewcommand\arraystretch{1.1}	\setlength{\tabcolsep}{3.0mm}
	\quad
	
	\begin{tabular}{l|l|ccc|c}
		\noalign{\hrule height 1pt}
		\multicolumn{2}{c|}{\textbf{Models}}                        &\ \ \ \ \ \textbf{IoU=0.3} \ \ \ \ \ & \ \ \ \ \ \textbf{IoU=0.5}\ \ \ \ \ &\ \ \ \ \ \textbf{IoU=0.7}\ \ \ \ \ &\ \ \ \ \ \textbf{mIoU} \ \ \ \ \ \\ \hline
		\multicolumn{2}{c|}{\textbf{Random Mode}}                      & 8.38&	1.93&	1.21 &	6.89     \\ 
		\multicolumn{2}{c|}{\textbf{Random Guess}}                      & 7.74    & 3.22            & 0.64            &   5.96\\ \hline
		\multirow{2}{*}{\rotatebox[origin=c]{0}{\textbf{VSLBase}\cite{HaoZhang2020SpanbasedLN} (2020)}}
		& with subtitle                &  27.66   & 14.19          & 6.99           & 21.01    \\    
		& w/o subtitle       & {26.12} & {12.44} & {6.85} &  {20.84}               \\ \hline
		\multirow{2}{*}{\rotatebox[origin=c]{0}{\textbf{TMLGA}\cite{opazo2019proposal} (2020)}}
		& with subtitle                     & 26.90   & 15.86            & 9.66            & 20.49\\    
		& w/o subtitle                         &  24.83   & 16.55           & 6.21            & 19.80       \\ \hline
		\multirow{2}{*}{\rotatebox[origin=c]{0}{\textbf{VSLNet}\cite{HaoZhang2020SpanbasedLN} (2020)}}
		& with subtitle                      &  33.10   & 16.61            & 8.39          & 22.61 \\ 		  
		& w/o subtitle                        & {30.32} & {16.55} & {7.74} &  {22.23}\\\hline
		\multirow{2}{*}{\rotatebox[origin=c]{0}{\textbf{VSLNet-L}\cite{HaoZhang2021NaturalLV} (2021)}}
		& with subtitle                      &  31.61   & 17.41            & 9.72          & 24.37 \\    
		& w/o subtitle                         &  29.03   & 16.77            & 9.03            & 23.09       \\ \hline
		\multirow{2}{*}{\rotatebox[origin=c]{0}{\textbf{ACRM}\cite{HaoyuTang2021FramewiseCM} (2021)}}
		& with subtitle                     &  26.90   & 18.06           & 12.90             & 23.70 \\    
		& w/o subtitle                         & 24.83   & 16.55            & 10.96             & 22.89       \\ \hline
		\multirow{2}{*}{\rotatebox[origin=c]{0}{\textbf{RaNet}\cite{HaoyuTang2021FramewiseCM} (2021)}}
		& with subtitle                      &  \underline{35.48}   & \underline{26.45}           &  14.84             & \underline{29.45} \\    
		& w/o subtitle                         & 32.90   & 20.64            & \underline{15.48}            & 27.48       \\ \hline
		\multicolumn{2}{c|}{\textbf{VPTSL}}            & \textbf{77.42}(+41.94$\uparrow$)  & \textbf{61.94}(+35.49$\uparrow$)           & \textbf{44.52}(+29.04$\uparrow$)           & \textbf{57.81 }(+28.36$\uparrow$)         \\ 		\noalign{\hrule height 1pt}
	\end{tabular}%
	
	\caption{Performance comparison of various SOTA methods on MedVidQA dataset. Here ``with subtitle'' means that the subtitle text in the video is added to the text features. We highlight the best score in each column in {bold}, and the second best score is marked with \underline{underline}. We also show the improvement between the proposed {VPTSL} and second place.}

\label{tab:medvidqa-results}
\vspace{-0.5cm}
\end{table*}

\subsection{Comparison with State-of-the-Art Methods}
We compare our VPTSL with several state-of-the-art (SOTA) methods on the MedVidQA dataset. Notably, we set the same I3D feature \cite{JoaoCarreira2017QuoVA} and text feature extraction model (i.e., DeBERTa pre-trained language model) as the visual and textual feature extractor respectively for all methods to ensure fairness. 

\textbf{TMLGA} \cite{opazo2019proposal} is the model with a dynamic filter, which adaptively transfers language information to visual domain attention map. A new loss function is designed to guide the model with the most relevant part of the video, and soft labels are performed to cope with annotation uncertainties.

\textbf{VSLBase} \cite{HaoZhang2020SpanbasedLN} is a standard span-based QA framework. Specifically, visual features are analogous to that of text passage, where the target moment is regarded as the answer span. The VSLBase is trained to predict the start and end times of the visual answer span. 

\textbf{VSLNet}  \cite{HaoZhang2020SpanbasedLN} introduces a Query-Guided Highlighting (QGH) strategy to further enhance the VSLBase model. The VSLNet regards the target moment and its adjacent contexts as foreground, while the rest as background, i.e., foreground covers a slightly longer span than the answer span.

\textbf{VSLNet-L}  \cite{HaoZhang2021NaturalLV} incorporates the concepts from multi-paragraph question answering \cite{wang2018multi} by applying a multi-scale split-and-concat-enation strategy to address the performance degradation on a long video. Long videos are segmented into multiple short clips. The hierarchical searching strategy is designed for more accurate moment localization.

\textbf{ACRM}  \cite{HaoyuTang2021FramewiseCM} predicts the temporal grounding based on an interaction modeling between vision and language modalities. Specifically, the attention module is introduced to automatically assign hidden features to query text with richer semantic information, which is considered to be more important for finding relevant video content. Moreover, the additional predictor is designed for utilizing the internal frames during training to improve the localization accuracy.

\textbf{RaNet}\cite{gao2021relation} represent the relation-aware network, which formulates temporal language grounding in the video inspired by reading comprehension \cite{du2017learning}.  The framework of RaNet is designed to select a grounding moment from the predefined answer collections with the aid of coarse-and-fine choice-query interaction and choice-choice relation construction. The choice-query interactor is proposed to match the visual and textual information simultaneously in sentence-moment and token-moment levels, leading to a coarse-and-fine cross-modal interaction. 
	\vspace{0.1cm}
\subsection{Implementation Details}
\begin{figure*}
	\centering
	\subfigure[Ablation study of hyper parameter $\alpha$.]{\includegraphics[width=8.2cm]{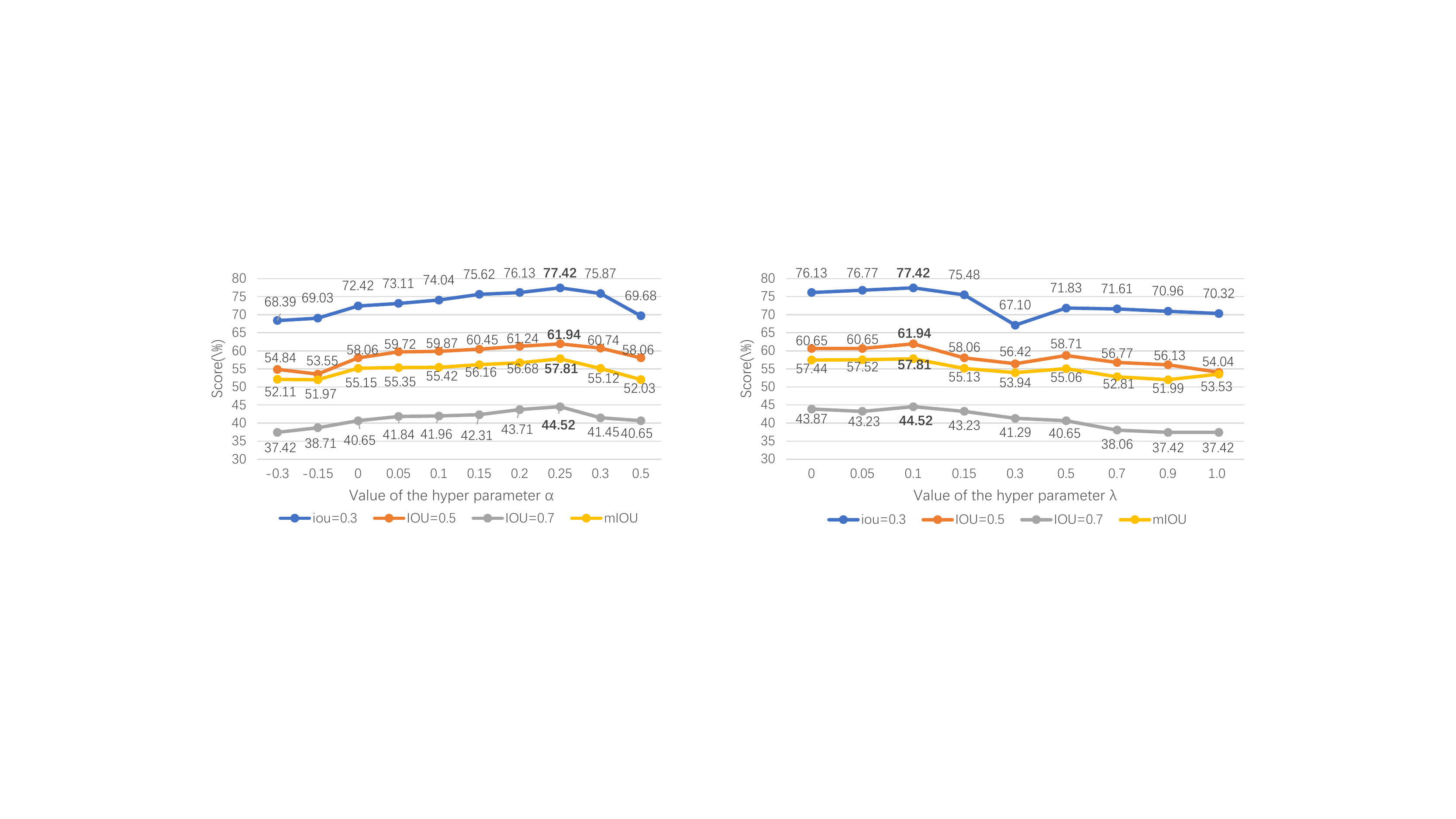} \label{s1}}
	\subfigure[Ablation study of hyper parameter $\lambda$.]{\includegraphics[width=9cm]{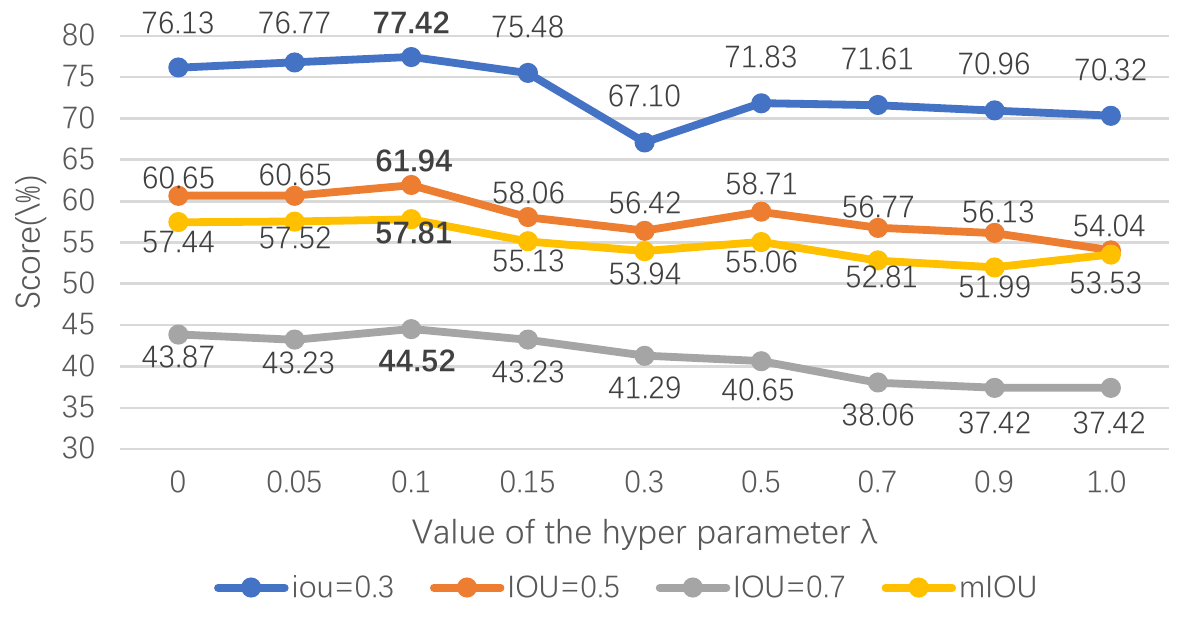}\label{s2}}
	\caption{Ablation study of hyper parameters $\alpha$ and $\lambda$ of the proposed VPTSL.}
	\vspace{-0.2cm}
\end{figure*}
We apply the same multimodal features for all the experiments for all the compared methods for fair comparisons. Specifically, for the textual features, we use the DeBERTa-v3 \cite{he2021debertav3} as the pre-trained language model, which originates from DeBERTa\footnote{\url{https://huggingface.co/microsoft/deberta-v3-large}} model with 24 layers and a hidden size of 1024. It has 304M backbone parameters and a vocabulary containing 128K tokens, introducing 131M parameters in the embedding layer. For visual features, all different methods adopt the I3D features \cite{JoaoCarreira2017QuoVA} as the visual input. We reproduce the compared method with the Pytorch\footnote{\url{https://pytorch.org}} \cite{NEURIPS2019_bdbca288} on three NVIDIA A100 GPUs, where all the implementations use the hugging-face\footnote{\url{https://github.com/huggingface/transformers}} \cite{wolf-etal-2020-transformers} framework. 
For the re-initiated layers, we set the dimension of all the hidden layers in the model as 1024 while the kernel size of convolution layers \cite{YoonKim2014ConvolutionalNN} is set to 7. The head size of multi-head attention \cite{10.5555/3295222.3295349} is 32. The text span predictor is initialized with another DeBERTa-v3 pre-trained language model, where the subtitles are essential to the proposed method. 
\par
As for the method adopting visual span predictor, we also compare their performances with the timestamped subtitles in addition to the original implementations. Specifically, the original implementations use the text question as the query to match the visual answer span for the TAGV task. To make use of the timestamped subtitles in these methods, we concatenated the text question and the subtitles with [SEP] separator, which are used to query the visual frames for cross-modal modeling. The start and end frames are obtained through the visual span predictor.
\par
We use the AdamW\cite{IlyaLoshchilov2018DecoupledWD} as the optimizer and the learning rate is set to 1e-5 with the warm-up \cite{7780459}. The batch size is 4. Moreover, we set the maximum length of 1800, and delete the excess part. The linear decay of learning rate and gradient clipping of 1e-6 and the dropout \cite{NitishSrivastava2014DropoutAS} is set to 0.1, which is applied to prevent overfitting. The hyperparameters of all the compared methods are tuned on the valid set. At the end of each training epoch, we test in the valid set and select the model with the highest score (mainly depending on mIOU) to predict in the test dataset. All the experimental implementations were repeated three times before reporting in the test set.
\section{Experimental Results}
\begin{table}[t]
	\label{tab:ha}
	\renewcommand\arraystretch{1.1}	\setlength{\tabcolsep}{1.4mm}
	\begin{tabular}{c|c|ccc|c}
		\noalign{\hrule height 1pt}
		\textbf{Method} &\textbf{Text Feature}  & \textbf{Iou=0.3}  & \textbf{Iou=0.5} & \textbf{Iou=0.7} & \textbf{mIOU} \\
		\hline
		\multirow{2}{*}{\rotatebox[origin=c]{0}{\textbf{VSLBase}}}
		 &Word2vec  &  21.93   & 12.25            & 5.80             & 20.15       \\ 
		&PLM  & \textbf{26.12} & \textbf{12.44} & \textbf{6.85} &  \textbf{20.84}  \\
				\hline
		\multirow{2}{*}{\rotatebox[origin=c]{0}{\textbf{VSLNet}}}
		&Word2vec  &  25.81   & 14.20            & 6.45             & 20.12  \\
		&PLM  & \textbf{30.32} & \textbf{16.55} & \textbf{7.74} &  \textbf{22.23}  \\
		\noalign{\hrule height 1pt}
	\end{tabular}
	\caption{Comparison of whether using the pre-trained language model (PLM) for VSLBase and VSLNet in MedVidQA}
	\label{PLM}
	\vspace{-0.3cm}
\end{table}

\begin{table}[t]
	
	\label{ta}
	\renewcommand\arraystretch{1.1}	\setlength{\tabcolsep}{1.9mm}
	\begin{tabular}{c|ccc|c}
		
		\noalign{\hrule height 1pt}
		\textbf{Experimental Item}  & \textbf{Iou=0.3}  & \textbf{Iou=0.5} & \textbf{Iou=0.7} & \textbf{mIOU} \\
		\hline
		W/o Highlight Loss & 76.13& 60.65 & 43.87 & 57.44 \\
		W/o Visual Prompt & 70.97 & 59.35 &43.87  &55.41  \\
		W/o PLM & 20.65 &10.97 &4.52  &18.83  \\
		\hline
		\textbf{VPTSL} & \textbf{77.42}  & \textbf{61.94}          & \textbf{44.52}           & \textbf{57.81} \\
		\noalign{\hrule height 1pt}
	\end{tabular}
	\caption{Results of ablation experiment on the MedVidQA dataset.}
	\label{PLM2}
	\vspace{-0.9cm}
\end{table}
\subsection{Main Results}
The experimental results of the performance compared with the various SOTA methods on the MedVidQA dataset are shown in Table \ref{tab:medvidqa-results}. The further conclusions are that our method outperforms each compared method on all metrics, including IOU= 0.3, 0.5, 0.7, and mIOU scores. The text span predictor can achieve better results than the method with the visual span predictor by a large margin in the mIOU score (28.36), indicating that the textual predictor is superior to the visual span predictor in locating the visual answer queried by the text question. The reason may be that the powerful pre-trained language model can leverage more strong semantics from the subtitles given the text question. Moreover, we also add the subtitles to each compared method adopting visual span predictor, where the text question with the subtitles of the video are concatenated with [SEP] for obtaining the text features. It can be found that the final results can be improved with the subtitles augmentation. However, the proposed VPTSL method still obtains significant improvements over these modified compared methods, which demonstrates the effectiveness of the visual prompt and the text span predictor.

\begin{figure*}[t]
	\centering
	\includegraphics[width=15.6cm]{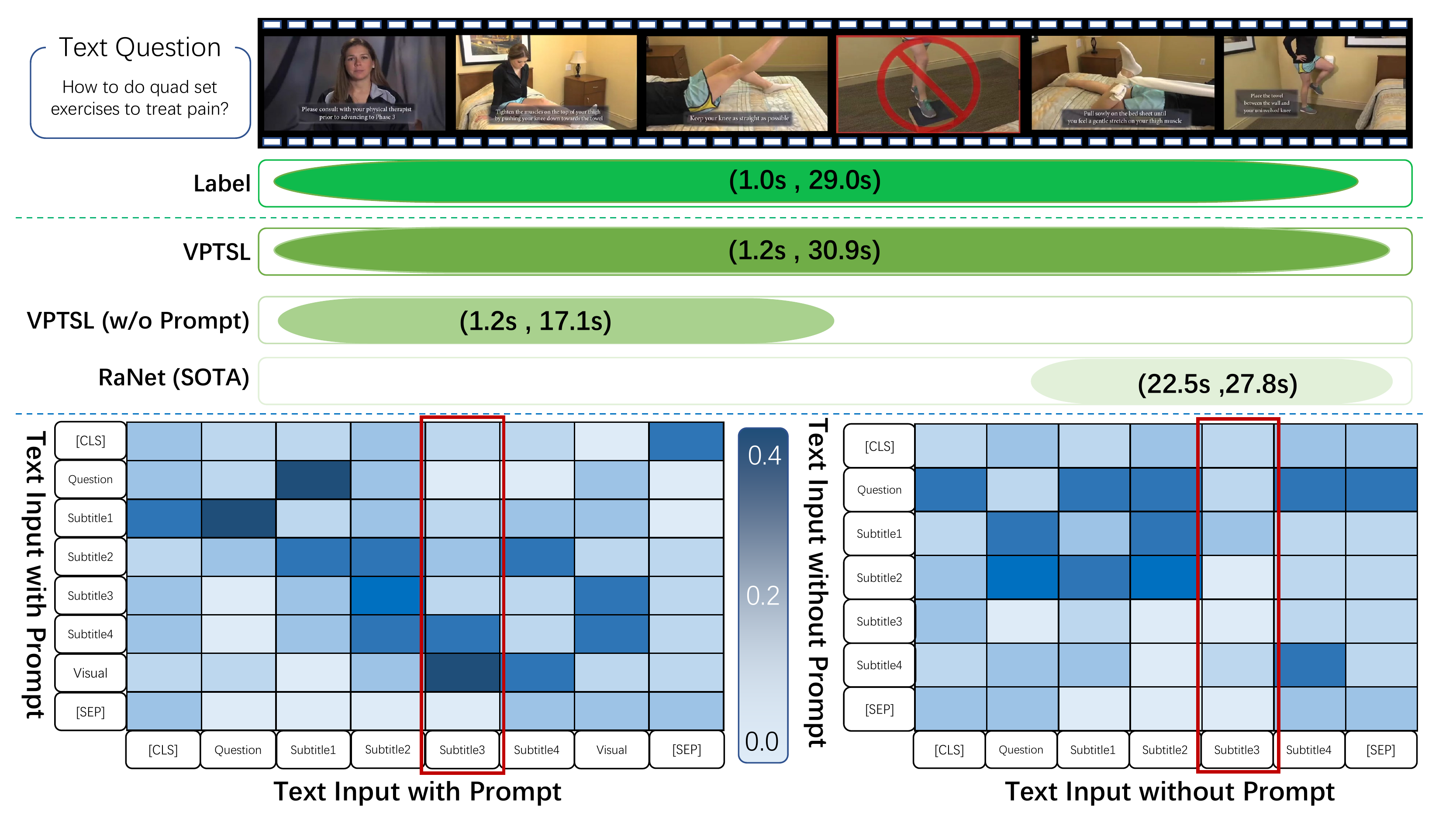}
	\caption{Case study of the proposed method compared with the SOTA method, where the embedding weights are visualized to demonstrate the visual prompt's effectiveness. The red box shows the semantics differences between the with and without visual prompt, where the visual prompt can gather more attention weights from the adjacent subtitles containing verbal and non-verbal parts for better video understanding.}
	\label{framework222}
	\vspace{-0.2cm}
\end{figure*}
	\vspace{-0.2cm}
\subsection{Ablation Studies}
We first investigate the effectiveness of the pre-trained model. The results of whether the classic visual span-based methods VSLBase and VSLNet adopt the pre-trained model are shown in Table~\ref{PLM}, where part of the results is in line with the work \cite{gupta2022dataset}. As we can see, only using the word2vec to initial the textual embedding layer achieves worse performance. In particular, the PLM for the textual feature extraction can improve semantic understanding, which results in improvements of 0.69 and 2.11 in terms of mIOU score for VSLBase and VSLNet respectively. It convincingly demonstrates that the pre-trained language model for textual feature extraction can improve the video understanding for the visual span-based method in this TAGV problem.
\par
Then we study the ablation of each component of the proposed method, which is shown in Table \ref{PLM2}. Specifically, for the w/o highlight loss, we remove the highlight loss from the total loss for training. For the w/o visual prompt, we use the question and timestamped subtitles to implement the text predictor for text span prediction. For w/o PLM, instead of loading pre-trained language model weights, we use a model of the same size and initialize the model parameters randomly. When there is no highlight supervision, the performance drops slightly. When the prompt module is removed, it is reduced by 2.4 mIOU compared to the original foundation. When the pre-trained language model is removed, the text span prediction ability is greatly impaired. So the performance improvement still comes from the strong understanding ability of the pre-trained model.
\par
We also analyze the performances under different hyperparameters in the experiment. As shown in Sub-Figure \ref{s1}, we show the performances with different values of the extend time hyperparameter $\alpha$ on the final result. It can be found from the ablation diagram that when the extend rate is 0.25, the best result is obtained. When $\alpha$ becomes larger, it decreases the performance of the final prediction result, which may be due to the non-visual part affecting the understanding of the input text. At the same time, when extend rate is less than 0, the text input range is insufficient, which results in bad performance.
\par
Meanwhile, we also study the weight of the visual prompt loss for the proposed method. From the Sub-Figure \ref{s2}, it can be found that the best results can be achieved when $\lambda$ is 0.1. Compared with no visual supervision, visual information can provide more contextual information for the prediction.
\subsection{Case Study}
We present the case study for the TAGV task shown in Figure \ref{framework222}. As we can see that, the proposed VPTSL has a better performance than the SOTA method (RaNet), where more precise subtitle spans are predicted by the text span predictor. Moreover, we compare the performance of whether to use prompt tuning in the prediction. It can be further concluded that the visual prompt can bridge the non-verbal part and the verbal part, bringing improvement to the final prediction.
\par
What's more, we also provide the visualization of the attention features in the embedding layer of the proposed method. Intuitively, these visualizations give the insight to dive into the function of the prompt. As shown in Figure \ref{framework222}, comparing the weights of subtitle 3 boxed in red with or without visual prompt, it receives more attention from its adjacent subtitles which are all in the answer span when the visual prompt is used.  This can be evidence that the visual prompt features can bridge the verbal part and the non-verbal part,  which results in more precise visual answer prediction.
\par

\section{Conclusion}
In this paper, we proposed the visual prompt text span localizing (VPTSL) method to make full use of complementary information of textual and visual semantic representations for temporal answering grounding in the video (TAGV). To this end, we first model the cross-modal information and proposed the visual prompt for enhancing the pre-trained language model. The text span predictor is designed for modeling the textual and visual representation for subtitle span prediction. The main results and ablation studies on the proposed method in the TAGV benchmarks significantly demonstrated the effectiveness of the VPTSL. In the future, the more precise and efficient method to perform the TAGV task with visual augmentation is yet to be explored.
\section*{Acknowledgement}
{This work is supported by the National Key R\&D Program of China (2018YFB1305200), the National Natural Science Fund of China (62171183).}
\bibliographystyle{unsrt}
\bibliography{sample-base}

\begin{thebibliography}{10}

\bibitem{monfort2021spoken}
Mathew Monfort, SouYoung Jin, Alexander Liu, David Harwath, Rogerio Feris,
  James Glass, and Aude Oliva.
\newblock Spoken moments: Learning joint audio-visual representations from
  video descriptions.
\newblock In {\em Proceedings of the IEEE/CVF Conference on Computer Vision and
  Pattern Recognition}, pages 14871--14881, 2021.

\bibitem{shelke2021comprehensive}
Nitin~Arvind Shelke and Singara~Singh Kasana.
\newblock A comprehensive survey on passive techniques for digital video
  forgery detection.
\newblock {\em Multimedia Tools and Applications}, 80(4):6247--6310, 2021.

\bibitem{khan2021transformers}
Salman Khan, Muzammal Naseer, Munawar Hayat, Syed~Waqas Zamir, Fahad~Shahbaz
  Khan, and Mubarak Shah.
\newblock Transformers in vision: A survey.
\newblock {\em ACM Computing Surveys (CSUR)}, 2021.

\bibitem{ghoddoosian2022hierarchical}
Reza Ghoddoosian, Saif Sayed, and Vassilis Athitsos.
\newblock Hierarchical modeling for task recognition and action segmentation in
  weakly-labeled instructional videos.
\newblock In {\em Proceedings of the IEEE/CVF Winter Conference on Applications
  of Computer Vision}, pages 1922--1932, 2022.

\bibitem{MedVidQA2022}
Deepak Gupta and Dina Demner-Fushman.
\newblock {Overview of the MedVidQA 2022 Shared Task on Medical Video Question
  Answering}.
\newblock In {\em Proceedings of the 21st SIGBioMed Workshop on Biomedical
  Language Processing, ACL-BioNLP 2022}. Association for Computational
  Linguistics, 2022.

\bibitem{gupta2022dataset}
Deepak Gupta, Kush Attal, and Dina Demner-Fushman.
\newblock {A Dataset for Medical Instructional Video Classification and
  Question Answering}.
\newblock {\em arXiv preprint arXiv:2201.12888}, 2022.

\bibitem{yang2020survey}
Yulan Yang, Zhaohui Li, and Gangyan Zeng.
\newblock A survey of temporal activity localization via language in untrimmed
  videos.
\newblock In {\em 2020 International Conference on Culture-oriented Science \&
  Technology (ICCST)}, pages 596--601. IEEE, 2020.

\bibitem{zhang2022elements}
Hao Zhang, Aixin Sun, Wei Jing, and Joey~Tianyi Zhou.
\newblock The elements of temporal sentence grounding in videos: A survey and
  future directions.
\newblock {\em arXiv preprint arXiv:2201.08071}, 2022.

\bibitem{chen2018temporally}
Jingyuan Chen, Xinpeng Chen, Lin Ma, Zequn Jie, and Tat-Seng Chua.
\newblock Temporally grounding natural sentence in video.
\newblock In {\em Proceedings of the 2018 conference on empirical methods in
  natural language processing}, pages 162--171, 2018.

\bibitem{zhang2020span}
Hao Zhang, Aixin Sun, Wei Jing, and Joey~Tianyi Zhou.
\newblock Span-based localizing network for natural language video
  localization.
\newblock In {\em Proceedings of the 58th Annual Meeting of the Association for
  Computational Linguistics}, pages 6543--6554, 2020.

\bibitem{xiao2021natural}
Shaoning Xiao, Long Chen, Jian Shao, Yueting Zhuang, and Jun Xiao.
\newblock Natural language video localization with learnable moment proposals.
\newblock {\em arXiv preprint arXiv:2109.10678}, 2021.

\bibitem{zhang2021natural}
Hao Zhang, Aixin Sun, Wei Jing, Liangli Zhen, Joey~Tianyi Zhou, and Rick
  Siow~Mong Goh.
\newblock Natural language video localization: A revisit in span-based question
  answering framework.
\newblock {\em IEEE Transactions on Pattern Analysis and Machine Intelligence},
  2021.

\bibitem{9374685}
H.~{Tang}, J.~{Zhu}, M.~{Liu}, Z.~{Gao}, and Z.~{Cheng}.
\newblock Frame-wise cross-modal matching for video moment retrieval.
\newblock {\em IEEE Transactions on Multimedia}, pages 1--1, 2021.

\bibitem{lei2018tvqa}
Jie Lei, Licheng Yu, Mohit Bansal, and Tamara Berg.
\newblock Tvqa: Localized, compositional video question answering.
\newblock In {\em Proceedings of the 2018 Conference on Empirical Methods in
  Natural Language Processing}, pages 1369--1379, 2018.

\bibitem{opazo2019proposal}
Cristian Rodríguez-Opazo, Edison Marrese-Taylor, Fatemeh~Sadat Saleh, Hongdong
  Li, and Stephen Gould.
\newblock Proposal-free temporal moment localization of a natural-language
  query in video using guided attention.
\newblock {\em Winter Conference on Applications of Computer Vision}, 2020.

\bibitem{HaoyuTang2021FramewiseCM}
Haoyu Tang, Jihua Zhu, Meng Liu, Zan Gao, and Zhiyong Cheng.
\newblock Frame-wise cross-modal matching for video moment retrieval.
\newblock {\em IEEE Transactions on Multimedia}, pages 1--1, 2021.

\bibitem{xu2022transformers}
Yifan Xu, Huapeng Wei, Minxuan Lin, Yingying Deng, Kekai Sheng, Mengdan Zhang,
  Fan Tang, Weiming Dong, Feiyue Huang, and Changsheng Xu.
\newblock Transformers in computational visual media: A survey.
\newblock {\em Computational Visual Media}, 8(1):33--62, 2022.

\bibitem{9262795}
Y.~{Yang}, Z.~{Li}, and G.~{Zeng}.
\newblock A survey of temporal activity localization via language in untrimmed
  videos.
\newblock In {\em 2020 International Conference on Culture-oriented Science
  Technology (ICCST)}, pages 596--601, 2020.

\bibitem{anne2017localizing}
Lisa Anne~Hendricks, Oliver Wang, Eli Shechtman, Josef Sivic, Trevor Darrell,
  and Bryan Russell.
\newblock Localizing moments in video with natural language.
\newblock In {\em Proceedings of the IEEE international conference on computer
  vision}, pages 5803--5812, 2017.

\bibitem{gao2017tall}
Jiyang Gao, Chen Sun, Zhenheng Yang, and Ram Nevatia.
\newblock Tall: Temporal activity localization via language query.
\newblock In {\em Proceedings of the IEEE international conference on computer
  vision}, pages 5267--5275, 2017.

\bibitem{MengLiu2018AttentiveMR}
Meng Liu, Xiang Wang, Liqiang Nie, Xiangnan He, Baoquan Chen, and Tat-Seng
  Chua.
\newblock Attentive moment retrieval in videos.
\newblock In {\em International ACM SIGIR Conference on Research and
  Development in Information Retrieval}, 2018.

\bibitem{chen-etal-2018-temporally}
Jingyuan Chen, Xinpeng Chen, Lin Ma, Zequn Jie, and Tat-Seng Chua.
\newblock Temporally grounding natural sentence in video.
\newblock In {\em Proceedings of the 2018 Conference on Empirical Methods in
  Natural Language Processing}, pages 162--171, Brussels, Belgium, 2018.
  Association for Computational Linguistics.

\bibitem{ge2019mac}
Runzhou Ge, Jiyang Gao, Kan Chen, and Ram Nevatia.
\newblock Mac: Mining activity concepts for language-based temporal
  localization.
\newblock In {\em 2019 IEEE Winter Conference on Applications of Computer
  Vision (WACV)}, pages 245--253. IEEE, 2019.

\bibitem{zhang2019man}
Da~Zhang, Xiyang Dai, Xin Wang, Yuan-Fang Wang, and Larry~S Davis.
\newblock Man: Moment alignment network for natural language moment retrieval
  via iterative graph adjustment.
\newblock In {\em Proceedings of the IEEE/CVF Conference on Computer Vision and
  Pattern Recognition}, pages 1247--1257, 2019.

\bibitem{liu2020jointly}
Daizong Liu, Xiaoye Qu, Xiao-Yang Liu, Jianfeng Dong, Pan Zhou, and Zichuan Xu.
\newblock Jointly cross-and self-modal graph attention network for query-based
  moment localization.
\newblock In {\em Proceedings of the 28th ACM International Conference on
  Multimedia}, pages 4070--4078, 2020.

\bibitem{HaoZhang2020SpanbasedLN}
Hao Zhang, Aixin Sun, Wei Jing, and Joey~Tianyi Zhou.
\newblock Span-based localizing network for natural language video
  localization.
\newblock {\em arXiv: Computation and Language}, 2020.

\bibitem{10.1145/3394171.3414053}
Xiaoye Qu, Pengwei Tang, Zhikang Zou, Yu~Cheng, Jianfeng Dong, Pan Zhou, and
  Zichuan Xu.
\newblock Fine-grained iterative attention network for temporal language
  localization in videos.
\newblock In {\em Proceedings of the 28th ACM International Conference on
  Multimedia}, MM '20, page 4280–4288, New York, NY, USA, 2020. Association
  for Computing Machinery.

\bibitem{10.1145/3394171.3413840}
Da~Cao, Yawen Zeng, Meng Liu, Xiangnan He, Meng Wang, and Zheng Qin.
\newblock Strong: Spatio-temporal reinforcement learning for cross-modal video
  moment localization.
\newblock In {\em Proceedings of the 28th ACM International Conference on
  Multimedia}, MM '20, page 4162–4170, New York, NY, USA, 2020. Association
  for Computing Machinery.

\bibitem{liu2021pre}
Pengfei Liu, Weizhe Yuan, Jinlan Fu, Zhengbao Jiang, Hiroaki Hayashi, and
  Graham Neubig.
\newblock Pre-train, prompt, and predict: A systematic survey of prompting
  methods in natural language processing.
\newblock {\em arXiv preprint arXiv:2107.13586}, 2021.

\bibitem{devlin2019bert}
Jacob Devlin, Ming-Wei Chang, Kenton Lee, and Kristina Toutanova.
\newblock Bert: Pre-training of deep bidirectional transformers for language
  understanding.
\newblock In {\em Proceedings of the 2019 Conference of the North American
  Chapter of the Association for Computational Linguistics: Human Language
  Technologies, Volume 1 (Long and Short Papers)}, pages 4171--4186, 2019.

\bibitem{radford2019language}
Alec Radford, Jeffrey Wu, Rewon Child, David Luan, Dario Amodei, Ilya
  Sutskever, et~al.
\newblock Language models are unsupervised multitask learners.
\newblock {\em OpenAI blog}, 1(8):9, 2019.

\bibitem{reynolds2021prompt}
Laria Reynolds and Kyle McDonell.
\newblock Prompt programming for large language models: Beyond the few-shot
  paradigm.
\newblock In {\em Extended Abstracts of the 2021 CHI Conference on Human
  Factors in Computing Systems}, pages 1--7, 2021.

\bibitem{wei2021pretrained}
Colin Wei, Sang~Michael Xie, and Tengyu Ma.
\newblock Why do pretrained language models help in downstream tasks? an
  analysis of head and prompt tuning.
\newblock {\em Advances in Neural Information Processing Systems}, 34, 2021.

\bibitem{lester2021power}
Brian Lester, Rami Al-Rfou, and Noah Constant.
\newblock The power of scale for parameter-efficient prompt tuning.
\newblock In {\em Proceedings of the 2021 Conference on Empirical Methods in
  Natural Language Processing}, pages 3045--3059, 2021.

\bibitem{yao2021cpt}
Yuan Yao, Ao~Zhang, Zhengyan Zhang, Zhiyuan Liu, Tat-Seng Chua, and Maosong
  Sun.
\newblock Cpt: Colorful prompt tuning for pre-trained vision-language models.
\newblock {\em arXiv preprint arXiv:2109.11797}, 2021.

\bibitem{rao2021denseclip}
Yongming Rao, Wenliang Zhao, Guangyi Chen, Yansong Tang, Zheng Zhu, Guan Huang,
  Jie Zhou, and Jiwen Lu.
\newblock Denseclip: Language-guided dense prediction with context-aware
  prompting.
\newblock {\em arXiv preprint arXiv:2112.01518}, 2021.

\bibitem{zhou2021learning}
Kaiyang Zhou, Jingkang Yang, Chen~Change Loy, and Ziwei Liu.
\newblock Learning to prompt for vision-language models.
\newblock {\em arXiv preprint arXiv:2109.01134}, 2021.

\bibitem{liu2022things}
Xiao Liu, Da~Yin, Yansong Feng, and Dongyan Zhao.
\newblock Things not written in text: Exploring spatial commonsense from visual
  signals.
\newblock {\em arXiv preprint arXiv:2203.08075}, 2022.

\bibitem{zhou2022conditional}
Kaiyang Zhou, Jingkang Yang, Chen~Change Loy, and Ziwei Liu.
\newblock Conditional prompt learning for vision-language models.
\newblock {\em arXiv preprint arXiv:2203.05557}, 2022.

\bibitem{JoaoCarreira2017QuoVA}
Joao Carreira and Andrew Zisserman.
\newblock Quo vadis, action recognition? a new model and the kinetics dataset.
\newblock In {\em Computer Vision and Pattern Recognition}, 2017.

\bibitem{ma2021end}
Pingchuan Ma, Stavros Petridis, and Maja Pantic.
\newblock End-to-end audio-visual speech recognition with conformers.
\newblock In {\em ICASSP 2021-2021 IEEE International Conference on Acoustics,
  Speech and Signal Processing (ICASSP)}, pages 7613--7617. IEEE, 2021.

\bibitem{he2021debertav3}
Pengcheng He, Jianfeng Gao, and Weizhu Chen.
\newblock Debertav3: Improving deberta using electra-style pre-training with
  gradient-disentangled embedding sharing, 2021.

\bibitem{bahdanau2014neural}
Dzmitry Bahdanau, Kyunghyun Cho, and Yoshua Bengio.
\newblock Neural machine translation by jointly learning to align and
  translate.
\newblock {\em In International Conference on Learning Representations}, 2015.

\bibitem{rajpurkar2016squad}
Pranav Rajpurkar, Jian Zhang, Konstantin Lopyrev, and Percy Liang.
\newblock Squad: 100,000+ questions for machine comprehension of text.
\newblock In {\em Proceedings of the 2016 Conference on Empirical Methods in
  Natural Language Processing}, pages 2383--2392, 2016.

\bibitem{PengchuanZhang2021VinVLRV}
Pengchuan Zhang, Xiujun Li, Xiaowei Hu, Jianwei Yang, Lei Zhang, Lijuan Wang,
  Yejin Choi, and Jianfeng Gao.
\newblock Vinvl: Revisiting visual representations in vision-language models.
\newblock {\em arXiv: Computer Vision and Pattern Recognition}, 2021.

\bibitem{LisaAnneHendricks2017LocalizingMI}
Lisa~Anne Hendricks, Oliver Wang, Eli Shechtman, Josef Sivic, Trevor Darrell,
  and Bryan Russell.
\newblock Localizing moments in video with natural language.
\newblock In {\em International Conference on Computer Vision}, 2017.

\bibitem{JiyangGao2017TALLTA}
Jiyang Gao, Chen Sun, Zhenheng Yang, and Ram Nevatia.
\newblock Tall: Temporal activity localization via language query.
\newblock In {\em International Conference on Computer Vision}, 2017.

\bibitem{YitianYuan2018ToFW}
Yitian Yuan, Tao Mei, and Wenwu Zhu.
\newblock To find where you talk: Temporal sentence localization in video with
  attention based location regression.
\newblock {\em arXiv: Computer Vision and Pattern Recognition}, 2018.

\bibitem{WenhuiWang2017GatedSN}
Wenhui Wang, Nan Yang, Furu Wei, Baobao Chang, and Ming Zhou.
\newblock Gated self-matching networks for reading comprehension and question
  answering.
\newblock In {\em Meeting of the Association for Computational Linguistics},
  2017.

\bibitem{MinjoonSeo2016BidirectionalAF}
Minjoon Seo, Aniruddha Kembhavi, Ali Farhadi, and Hannaneh Hajishirzi.
\newblock Bidirectional attention flow for machine comprehension.
\newblock {\em arXiv: Computation and Language}, 2016.

\bibitem{HaoZhang2021NaturalLV}
Hao Zhang, Aixin Sun, Wei Jing, Liangli Zhen, Joey~Tianyi Zhou, and Rick
  Siow~Mong Goh.
\newblock Natural language video localization: A revisit in span-based question
  answering framework.
\newblock {\em IEEE Transactions on Pattern Analysis and Machine Intelligence},
  pages 1--1, 2021.

\bibitem{wang2018multi}
Wei Wang, Ming Yan, and Chen Wu.
\newblock Multi-granularity hierarchical attention fusion networks for reading
  comprehension and question answering.
\newblock In {\em Proceedings of the 56th Annual Meeting of the Association for
  Computational Linguistics (Volume 1: Long Papers)}, pages 1705--1714, 2018.

\bibitem{gao2021relation}
Jialin Gao, Xin Sun, Mengmeng Xu, Xi~Zhou, and Bernard Ghanem.
\newblock Relation-aware video reading comprehension for temporal language
  grounding.
\newblock {\em arXiv preprint arXiv:2110.05717}, 2021.

\bibitem{du2017learning}
Xinya Du, Junru Shao, and Claire Cardie.
\newblock Learning to ask: Neural question generation for reading
  comprehension.
\newblock In {\em Proceedings of the 55th Annual Meeting of the Association for
  Computational Linguistics (Volume 1: Long Papers)}, pages 1342--1352, 2017.

\bibitem{NEURIPS2019_bdbca288}
Adam Paszke, Sam Gross, Francisco Massa, Adam Lerer, James Bradbury, Gregory
  Chanan, Trevor Killeen, Zeming Lin, Natalia Gimelshein, Luca Antiga, Alban
  Desmaison, Andreas Kopf, Edward Yang, Zachary DeVito, Martin Raison, Alykhan
  Tejani, Sasank Chilamkurthy, Benoit Steiner, Lu~Fang, Junjie Bai, and Soumith
  Chintala.
\newblock Pytorch: An imperative style, high-performance deep learning library.
\newblock In H.~Wallach, H.~Larochelle, A.~Beygelzimer, F.~d\textquotesingle
  Alch\'{e}-Buc, E.~Fox, and R.~Garnett, editors, {\em Advances in Neural
  Information Processing Systems}, volume~32. Curran Associates, Inc., 2019.

\bibitem{wolf-etal-2020-transformers}
Thomas Wolf, Lysandre Debut, Victor Sanh, Julien Chaumond, Clement Delangue,
  Anthony Moi, Pierric Cistac, Tim Rault, Rémi Louf, Morgan Funtowicz, Joe
  Davison, Sam Shleifer, Patrick von Platen, Clara Ma, Yacine Jernite, Julien
  Plu, Canwen Xu, Teven~Le Scao, Sylvain Gugger, Mariama Drame, Quentin Lhoest,
  and Alexander~M. Rush.
\newblock Transformers: State-of-the-art natural language processing.
\newblock In {\em Proceedings of the 2020 Conference on Empirical Methods in
  Natural Language Processing: System Demonstrations}, pages 38--45, Online,
  October 2020. Association for Computational Linguistics.

\bibitem{YoonKim2014ConvolutionalNN}
Yoon Kim.
\newblock Convolutional neural networks for sentence classification.
\newblock In {\em Empirical Methods in Natural Language Processing}, 2014.

\bibitem{10.5555/3295222.3295349}
Ashish Vaswani, Noam Shazeer, Niki Parmar, Jakob Uszkoreit, Llion Jones,
  Aidan~N. Gomez, \L{}ukasz Kaiser, and Illia Polosukhin.
\newblock Attention is all you need.
\newblock In {\em Proceedings of the 31st International Conference on Neural
  Information Processing Systems}, NIPS'17, page 6000–6010, Red Hook, NY,
  USA, 2017. Curran Associates Inc.

\bibitem{IlyaLoshchilov2018DecoupledWD}
Ilya Loshchilov and Frank Hutter.
\newblock Decoupled weight decay regularization.
\newblock In {\em International Conference on Learning Representations}, 2018.

\bibitem{7780459}
Kaiming He, Xiangyu Zhang, Shaoqing Ren, and Jian Sun.
\newblock Deep residual learning for image recognition.
\newblock In {\em 2016 IEEE Conference on Computer Vision and Pattern
  Recognition (CVPR)}, pages 770--778, 2016.

\bibitem{NitishSrivastava2014DropoutAS}
Nitish Srivastava, Geoffrey~E. Hinton, Alex Krizhevsky, Ilya Sutskever, and
  Ruslan Salakhutdinov.
\newblock Dropout: a simple way to prevent neural networks from overfitting.
\newblock {\em Journal of Machine Learning Research}, 15:1929--1958, 2014.

\end{thebibliography}

\end{document}